\documentclass{article}
\usepackage{iclr2027_conference,times}
\usepackage[T1]{fontenc}
\usepackage[utf8]{inputenc}
\usepackage{amsmath,amssymb,booktabs,longtable,array,graphicx}
\usepackage{placeins}
\usepackage[colorlinks=true,citecolor=blue,linkcolor=blue,urlcolor=blue]{hyperref}
\ifdefined\XeTeXversion
  \AddToHook{cmd/@fancyhead/before}{}
  \AddToHook{cmd/@fancyhead/after}{}
  \AddToHook{cmd/@fancyfoot/before}{}
  \AddToHook{cmd/@fancyfoot/after}{}
  \AddToHook{cmd/iclrruler/before}{}
  \AddToHook{cmd/iclrruler/after}{}
\fi
\usepackage{url}
\usepackage{enumitem}
\title{TRACE: Trajectory Selection for Parallel Scaling of Search Agents}
\author{
Qisheng Zhou$^{1}$ \qquad Zhen Xiong$^{2}$ \qquad Qiaoyu Tan$^{1}$ \\
\normalfont $^{1}$New York University Shanghai \quad $^{2}$New York University
}
\iclrfinalcopy
\date{}
\hypersetup{
  pdftitle={TRACE: Trajectory Selection for Parallel Scaling of Search Agents},
  pdfauthor={Qisheng Zhou, Zhen Xiong, Qiaoyu Tan},
  pdfsubject={Trajectory selection for parallel scaling of search agents},
  pdfkeywords={search agents, trajectory selection, parallel scaling, graph neural networks}
}

\begin{document}
\maketitle
\fancyhead{}
\renewcommand{\headrulewidth}{0pt}
\begin{abstract}
Parallel scaling improves search agents by generating multiple candidate rollouts for the same query, yet a correct answer may already be present in the pool and still be missed by final-answer voting. We formulate this consolidation stage as \emph{trajectory selection} and introduce \textsc{TRACE} (Trajectory Ranking with Aggregated Cross-Rollout Evidence), a lightweight learned selector that ranks completed trajectories using the search evidence behind their answers. \textsc{TRACE} preserves individual query and evidence occurrences, connects rollouts through shared content or document identity, and propagates information across these relations. Each candidate answer then reads the updated states of its own trajectory, preserving retrieval provenance while incorporating evidence from related rollouts. Trained with answer-level supervision over frozen text embeddings, \textsc{TRACE} returns an existing answer without additional search or autoregressive aggregation. One selector per search setting transfers across rollout policies and agent backbones without agent-specific fine-tuning, improving over voting across six WebQA policies and six long-horizon dataset--backbone combinations at $K=16$. On Qwen2.5-14B Base/SFT WebQA pools, \textsc{TRACE} achieves 45.2/49.2\% EM, compared with 43.9/48.0\% for the strongest Qwen3-32B generative aggregators. On long-horizon FRAMES, GAIA, and BrowseComp, it reaches 78.6\% average accuracy, exceeding majority voting by 3.1 percentage points. On Base WebQA pools, \textsc{TRACE} with only 8 rollouts comes within 0.4 points of majority voting over 64. Beyond accuracy, \textsc{TRACE} is substantially more efficient than the three LLM-based aggregators, SolAgg, SummAgg, and AggAgent, achieving at least $10\times$ higher processing throughput across all seven WebQA benchmarks. These results show that reusing cross-rollout search evidence provides an effective and efficient alternative to heavyweight generative aggregation for parallel search. Code is available at \url{https://github.com/Jaasssoooonnnnn/TRACE}.

\end{abstract}

\section{Introduction}
\label{sec:introduction}
Search agents access external information through interleaved reasoning, search, and observation~\citep{yao2023react,jin2025searchr1}. A natural way to improve their test-time performance is \emph{parallel scaling}: for the same question, sample multiple completed search trajectories that explore different queries, sources, reasoning paths, and candidate answers~\citep{brown2024monkeys}. This strategy can substantially increase the chance of obtaining a successful solution, but it also introduces a distinct consolidation problem: \emph{given $K$ completed search rollouts, which trajectory should the system ultimately return?}

Existing approaches exploit the candidate pool in different ways. Answer-level voting selects by agreement among final outputs~\citep{wang2023selfconsistency}, but discards most of the search process that produced them. Learned verifiers score candidate solutions from their answers or reasoning traces~\citep{cobbe2021verifiers,montgomery2025budget}, yet typically evaluate candidates independently. More recent generative aggregators inspect multiple completed trajectories and invoke another large language model to synthesize a final response~\citep{lee2026aggagent}. While such methods can combine information across rollouts, they introduce an additional autoregressive reasoning stage after the expensive search trajectories have already been generated. When a correct answer is already present in the candidate pool, this extra generation may be unnecessary and can introduce new synthesis or formatting errors.

We instead formulate post-rollout consolidation as \emph{trajectory selection}: directly score the $K$ completed trajectories and return one of the existing candidates. This formulation is motivated by a simple observation: search trajectories contain information that is absent from their final answers alone~\citep{lee2026aggagent}. Independent rollouts may retrieve the same passage through different subqueries, or visit the same underlying source while exposing different passages (Appendix~\ref{app:graph_statistics}). These shared-evidence relations provide useful signals for assessing candidate trajectories. At the same time, naively merging repeated observations would erase the query and retrieval context in which each occurrence was obtained. Effective trajectory selection therefore requires a representation that can exchange information across rollouts while preserving the provenance of each search process.

To this end, we introduce \textsc{TRACE} (Trajectory Ranking with Aggregated Cross-Rollout Evidence), a lightweight selector that jointly evaluates completed search trajectories. \textsc{TRACE} constructs an occurrence-preserving query--evidence graph over the candidate pool. Within each rollout, it retains individual search and evidence occurrences together with their retrieval structure; across rollouts, it connects observations through shared underlying evidence---shared chunk identity for retrieval-based WebQA and shared document identity for long-horizon online search. A relation-specific graph encoder propagates information through these cross-rollout connections. Each candidate answer then attends only to the updated search and evidence states of its own trajectory, yielding a representation that remains locally grounded in the process that produced the answer while being globally informed by related searches. \textsc{TRACE} finally scores these candidate representations and returns the highest-scoring existing answer without additional search or autoregressive answer synthesis.

An important property of this design is that trajectory selection is decoupled from trajectory generation. \textsc{TRACE} is trained from answer-level correctness supervision over frozen text embeddings and does not modify the underlying search agent. One selector per search setting can therefore be applied directly to candidate pools generated by different rollout policies and agent backbones without target-generator fine-tuning. We evaluate this property across six WebQA rollout policies and long-horizon BrowseComp-Plus~\citep{chen2025browsecompplus}, FRAMES~\citep{krishna2024frames}, and GAIA~\citep{mialon2023gaia} benchmarks with different browsing agents. At $K=16$, \textsc{TRACE} consistently improves over answer-level voting and remains competitive with or stronger than Qwen3-32B generative aggregation. On Qwen2.5-14B Base/SFT WebQA pools, \textsc{TRACE} achieves 45.2/49.2\% EM, compared with 43.9/48.0\% for the strongest generative aggregators. On long-horizon search, it reaches 78.6\% average accuracy, exceeding majority voting by 3.1 percentage points. Moreover, on Base WebQA pools, selecting from only 8 trajectories comes within 0.4 points of majority voting over 64, showing that better post-rollout selection can substantially reduce the sampling budget required for competitive final-answer accuracy.
Our main \textbf{contributions} are summarized as follows:
\begin{itemize}[noitemsep,leftmargin=*]
\item \textbf{Trajectory selection for parallel scaling.}
We formulate post-rollout consolidation as a dedicated trajectory-selection problem: given $K$ completed search rollouts, identify which existing trajectory should be returned. This separates trajectory generation from final consolidation and provides an alternative to both answer-level voting and additional generative aggregation.

\item \textbf{Occurrence-preserving cross-rollout evidence reasoning.}
We introduce \textsc{TRACE}, which preserves each query and evidence occurrence while connecting trajectories through shared underlying evidence. Cross-rollout graph propagation followed by answer-conditioned readout yields candidate representations that retain their own retrieval context while incorporating information from related searches.

\item \textbf{Transferable and efficient selection across rollout generators.}
We show that one selector per search setting generalizes across six WebQA rollout policies and multiple long-horizon search-agent backbones without generator-specific retraining, consistently improving over voting and large-LLM aggregation baselines. We further show that learned selection reduces the rollout budget needed for competitive accuracy and replaces autoregressive aggregation with lightweight text encoding and graph inference.
\end{itemize}

\section{Related Work}
\label{sec:related_work}
\paragraph{Parallel scaling and trajectory aggregation.}
Test-time scaling improves language models through additional inference compute, including repeated sampling and parallel exploration~\citep{brown2024monkeys,snell2024scaling}. For search agents, recent work extends this paradigm to multiple tool-augmented trajectories~\citep{zhu2025agents,zeng2025asymmetric,li2025parallelmuse}. Existing consolidation strategies range from answer-level voting~\citep{wang2023selfconsistency} to generative aggregation over completed trajectories. In particular, AggAgent uses another language-model agent to inspect parallel rollouts and synthesize a final answer~\citep{lee2026aggagent}, while ParallelMuse reuses information across parallel search paths~\citep{li2025parallelmuse}. \textsc{TRACE} instead formulates consolidation as \emph{trajectory selection}: it exploits information across completed rollouts but returns one existing candidate rather than invoking an additional autoregressive generation stage.

\paragraph{Verification and candidate selection.}
Best-of-$N$ methods and learned verifiers select among sampled candidates using outcome-level or process-level supervision~\citep{cobbe2021verifiers,lightman2023verify,montgomery2025budget}. Most existing verifiers score candidates independently from their answer or reasoning trace. \textsc{TRACE} differs by treating trajectory quality as relational: the score of one rollout can depend on evidence encountered by other rollouts, which is especially relevant for search agents that may revisit the same chunk or source through different search paths.

\paragraph{Graph-structured evidence reasoning.}
Graphs have been used to organize evidence and reasoning dependencies in multi-hop QA, fact verification, and LLM reasoning~\citep{fang2020hierarchical,zhou2019gear,cao2024graphreason,hao2026gnnverifier}. \citet{xiong2025mapping} construct directed graphs of semantically clustered chain-of-thought steps to analyze how reasoning structure relates to answer accuracy. \textsc{TRACE} instead constructs an occurrence-preserving graph over a \emph{set of completed search trajectories}. It preserves each query and evidence occurrence while connecting rollouts through shared chunk or document identity, enabling cross-rollout message passing without erasing retrieval provenance before answer-conditioned trajectory scoring.

\section{TRACE: Trajectory Selection with Cross-Rollout Evidence}
\label{sec:method}
\subsection{Problem Setup and Overview}
\label{sec:method_overview}

Given a question $q$ and $K$ completed search trajectories
$\mathcal{T}_K=(\tau_1,\ldots,\tau_K)$, our goal is to select one existing trajectory rather than generate another answer.
Each trajectory $\tau_i$ contains its search process, retrieved evidence, and final answer $a_i$.
After validity filtering, let $\mathcal{I}_q$ index the retained candidates and
$R_q=|\mathcal{I}_q|\leq K$.
A selector $S$ returns
\begin{equation}
    \hat{\imath}=S(q,\mathcal{T}_K),
    \qquad
    \hat{a}=a_{\hat{\imath}},
    \label{eq:selection_problem}
\end{equation}
where $\hat{\imath}\in\mathcal{I}_q$. \textsc{TRACE} is designed around the observation that trajectory quality can depend on evidence encountered elsewhere in the candidate pool.
Different rollouts may retrieve the same chunk through different subqueries or inspect different passages from the same underlying document.
At the same time, collapsing these repeated observations would erase the query and retrieval context in which each occurrence was obtained.
\textsc{TRACE} therefore preserves individual search and evidence occurrences while explicitly connecting occurrences that share the same underlying evidence.

\begin{figure}[t]
    \centering
    \includegraphics[width=\linewidth]{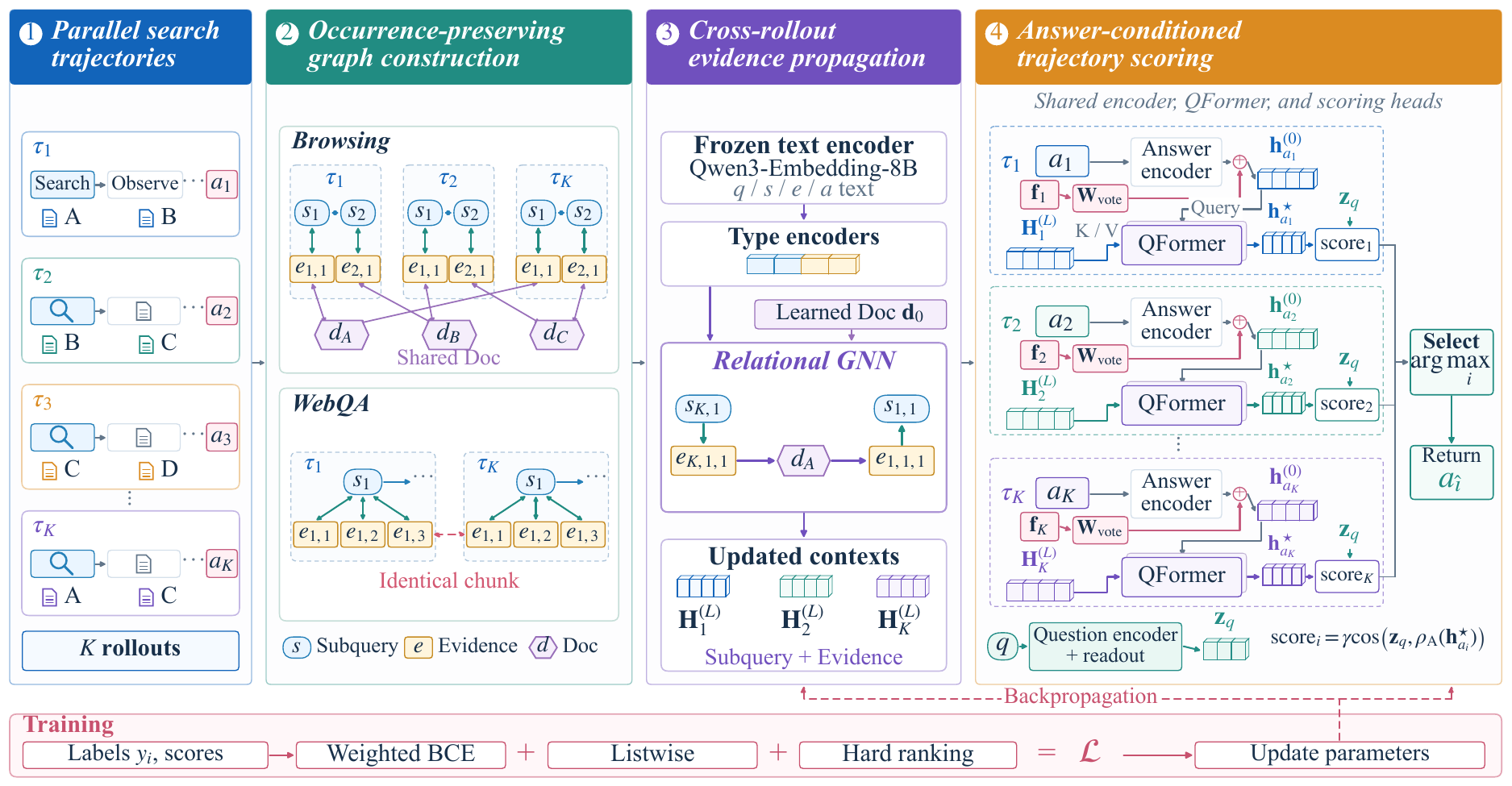}
    \caption{\textsc{TRACE} selects one answer from $K$ completed search trajectories.
    It first constructs an occurrence-preserving query--evidence graph, connecting trajectories through identical chunks in WebQA or shared documents in long-horizon browsing.
    A relation-specific GNN propagates information across these links.
    Each candidate answer then queries the updated Subquery/Evidence states of its own trajectory through a shared answer-conditioned QFormer and receives a selection score.
    The highest-scoring existing answer is returned without additional search or autoregressive generation.
    Graphs are schematic; WebQA's Query node and some within-rollout edges are omitted for clarity.}
    \label{fig:overview}
\end{figure}

Figure~\ref{fig:overview} summarizes the workflow.
\textsc{TRACE} first converts the completed trajectory pool into an occurrence-preserving query--evidence graph.
A relational GNN then propagates information across within-trajectory search relations and cross-rollout shared-evidence relations.
Finally, each candidate answer queries only the updated process states of its own trajectory through an answer-conditioned readout and receives a selection score.
Thus, cross-rollout reasoning occurs at the evidence level, while the final decision remains trajectory-specific.

\subsection{Occurrence-Preserving Query--Evidence Graph}
\label{sec:graph_construction}

We parse each completed interaction into search queries, returned evidence, source references, and a final answer.
Nodes correspond to concrete \emph{occurrences}: two searches that retrieve the same text still produce distinct Evidence nodes.
This choice preserves the retrieval provenance of each observation, including which query produced it and where it appeared in the trajectory.
Shared evidence is represented through graph relations rather than node merging.

Across both search protocols, let $s_{i,t}$ denote the $t$-th Subquery, or search-context, node in trajectory $i$, and let $e_{i,t,j}$ denote its $j$-th Evidence occurrence.
For WebQA, an Evidence node represents a returned text chunk.
For long-horizon browsing, it represents an observation produced by Open or Find.
We additionally introduce Doc nodes $d_m$ in the browsing setting to represent underlying document identity.

\paragraph{WebQA: ranked chunk retrieval.}
The agent alternates reasoning with Search calls, each returning three ranked text chunks before a final answer is produced.
We create one Subquery occurrence for every Search with a valid return and one Evidence occurrence for each returned chunk at ranks 1--3.
Rank-specific relations connect each Subquery to its retrieved chunks in both directions, while temporal edges preserve Search order within a rollout.
A Query node connects to the first Subquery of every trajectory.

When two Evidence occurrences contain the same normalized chunk, they are connected bidirectionally.
We distinguish within-rollout and cross-rollout matches with separate relation types.
Hence, repeated chunks can exchange search-context information while retaining their own rank, query association, and trajectory membership.

\paragraph{Long-horizon browsing: source-linked observations.}
Long-horizon agents interact with the web through Search, Open, and Find.
Every Search becomes a Subquery node, and each associated Open/Find observation becomes an Evidence occurrence connected to its originating Subquery through tool-specific bidirectional relations.
Successive Subqueries are linked by forward and backward temporal edges.
Appendix~\ref{app:data_protocols} specifies provenance tracking, including observations that do not have an explicit Search ancestor.

Exact-content matching is insufficient in this setting because different visits to the same source may expose different passages.
We therefore introduce a shared Doc node $d_m$ for each document identity and connect all Evidence occurrences from that source to it:
\begin{equation}
    e_{i,t,j}
    \leftrightarrow
    d_m
    \leftrightarrow
    e_{i',t',j'}.
    \label{eq:doc_relation}
\end{equation}
Cross-rollout information consequently travels through a two-hop
Evidence--Doc--Evidence path without merging the underlying observations.
Most same-document cross-rollout pairs in our analyzed browsing pools have different content identities (Appendix~\ref{app:graph_statistics}), supporting document identity as a useful relation beyond exact-text matching.
Appendix~\ref{app:additional_ablations} compares document- and content-based connection schemes.

\subsection{Cross-Rollout Evidence Propagation}
\label{sec:process_encoding}

Frozen Qwen3-Embedding-8B~\citep{zhang2025qwen3embedding} produces
4,096-dimensional embeddings $\mathbf{x}_v$ for questions, Subqueries, Evidence nodes, and answers.
Trainable type-specific encoders map them to
$\mathbf{h}_v^{(0)}\in\mathbb{R}^{d}$ with $d=256$.
For candidate answer $a_i$, the initial state additionally includes an answer-frequency feature:
\begin{equation}
    \mathbf{h}_{a_i}^{(0)}
    =
    \phi_{\mathrm{A}}(\mathbf{x}_{a_i})
    +
    \mathbf{W}_{\mathrm{vote}}\mathbf{f}_i,
    \qquad
    \mathbf{f}_i
    =
    \left[
        \log(1+c_i),\;
        \frac{c_i}{R_q}
    \right]^{\top},
    \label{eq:answer_initialization}
\end{equation}
where $c_i$ is the number of retained candidates with the same normalized answer.
Doc nodes share a learned initial vector $\mathbf{d}_0$, initialized to zero, and obtain document-specific information from their neighboring Evidence nodes.

We use $L=4$ relation-specific GraphSAGE layers~\citep{hamilton2017inductive}.
For node $v$ at layer $\ell$, messages are aggregated separately over relation types:
\begin{equation}
\begin{aligned}
    \mathbf{m}_v^{(\ell)}
    &=
    \sum_r
    \left(
        \mathbf{W}_r^{(\ell)}
        \operatorname{Mean}_{u\in\mathcal{N}_r(v)}
        \mathbf{h}_u^{(\ell)}
        +
        \mathbf{b}_r^{(\ell)}
    \right),\\
    \mathbf{h}_v^{(\ell+1)}
    &=
    \operatorname{LN}
    \left(
        \mathbf{h}_v^{(\ell)}
        +
        \operatorname{Dropout}
        \left(
            \operatorname{GELU}
            (\mathbf{m}_v^{(\ell)})
        \right)
    \right).
\end{aligned}
\label{eq:process_message_passing}
\end{equation}

Subquery and Evidence states are updated in both settings, together with Doc states in long-horizon browsing.
Query and Answer states receive no graph messages.
Thus, the graph encoder performs \emph{answer-independent evidence propagation}: matching WebQA chunks exchange information directly in one hop, whereas browsing observations communicate through their shared Doc node in two hops.
The complete process-encoding details are provided in Appendix~\ref{app:encoder_details}.

\subsection{Answer-Conditioned Trajectory Readout}
\label{sec:answer_readout}

After graph propagation, each candidate is evaluated from the updated process states of its own trajectory.
Let
\begin{equation}
    \mathcal{C}_i
    =
    \{
        v:
        v \text{ is a Subquery or Evidence occurrence in } \tau_i
    \},
\end{equation}
and let
$\mathbf{H}_i=[\mathbf{h}_v^{(L)}]_{v\in\mathcal{C}_i}$.
Although $\mathcal{C}_i$ contains only nodes from trajectory $i$, these states may already contain information propagated from other rollouts through shared-evidence relations.
This design therefore preserves local retrieval provenance while making each trajectory globally informed.

We use the candidate answer as the query of a shared cross-attention block, which we term an answer-conditioned QFormer.
A learned type embedding is added to each process state:
\begin{equation}
    \widetilde{\mathbf{h}}_v
    =
    \mathbf{h}_v^{(L)}
    +
    \mathbf{u}_{\operatorname{type}(v)}.
\end{equation}
For one attention head of width $d_{\mathrm{h}}$,
\begin{equation}
    \alpha_{i,v}
    =
    \operatorname{softmax}_{v\in\mathcal{C}_i}
    \left(
        \frac{
            (\mathbf{W}_{\mathrm{Q}}\mathbf{h}_{a_i}^{(0)})^{\top}
            (\mathbf{W}_{\mathrm{K}}\widetilde{\mathbf{h}}_v)
        }{
            \sqrt{d_{\mathrm{h}}}
        }
    \right),
    \qquad
    \mathbf{z}_i
    =
    \sum_{v\in\mathcal{C}_i}
    \alpha_{i,v}
    \mathbf{W}_{\mathrm{V}}
    \widetilde{\mathbf{h}}_v.
    \label{eq:answer_attention}
\end{equation}
Four attention heads followed by residual and feed-forward updates produce the candidate representation
$\mathbf{h}_{a_i}^{\star}$.

Importantly, answer states do not participate in graph message passing.
Candidate conditioning is introduced only at this readout stage.
The computation therefore separates
\emph{cross-rollout evidence propagation}
from
\emph{candidate-specific verification}:
\begin{equation}
    \mathbf{h}_{a_i}^{\star}
    =
    f(a_i,\tau_i,\mathcal{T}_K).
    \label{eq:global_local_representation}
\end{equation}
Each representation remains locally grounded in trajectory $\tau_i$, while indirectly incorporating evidence from related rollouts.

Finally, trainable projections $\rho_{\mathrm{Q}}$ and $\rho_{\mathrm{A}}$ map the question and updated answer into a common scoring space.
Let
$\mathbf{z}_q=\rho_{\mathrm{Q}}(\mathbf{h}_q^{(0)})$.
We compute
\begin{equation}
    \mathrm{score}_i
    =
    \gamma
    \cos
    \left(
        \mathbf{z}_q,
        \rho_{\mathrm{A}}(\mathbf{h}_{a_i}^{\star})
    \right),
    \qquad
    \hat{\imath}
    =
    \arg\max_{i\in\mathcal{I}_q}
    \mathrm{score}_i,
    \qquad
    \hat{a}
    =
    a_{\hat{\imath}},
    \label{eq:selector_score}
\end{equation}
where $\gamma$ is learned.
Identical answer strings may receive different scores because they remain associated with distinct search trajectories and process contexts.
The highest-scoring existing candidate is returned directly.

\subsection{Learning to Select}
\label{sec:selector_training}

Training uses answer-level correctness labels
$\mathbf{y}_q=(y_i)_{i\in\mathcal{I}_q}$,
where $y_i\in\{0,1\}$ is obtained by exact match for WebQA or the available correctness annotations for long-horizon browsing.
Let
\begin{equation}
    \mathcal{P}_q
    =
    \{i\in\mathcal{I}_q:y_i=1\},
    \qquad
    \mathcal{N}_q
    =
    \mathcal{I}_q\setminus\mathcal{P}_q.
\end{equation}

We optimize three complementary objectives.
Weighted binary cross-entropy learns candidate-level correctness;
a listwise objective concentrates probability mass on correct trajectories within the candidate pool;
and a hard-ranking objective separates the highest-scoring incorrect candidate from the strongest correct candidate.
The full objective is
\begin{equation}
    \mathcal{L}
    =
    \mathcal{L}_{\mathrm{BCE}}
    +
    \mathcal{L}_{\mathrm{list}}
    +
    \mathcal{L}_{\mathrm{hard}}.
    \label{eq:training_objective}
\end{equation}
BCE is averaged over candidate nodes, while the ranking terms are computed only for questions for which their required positive/negative sets are defined.
Appendix~\ref{app:encoder_details} provides the exact loss definitions and eligibility conditions.

Gradients jointly update the type-specific encoders, relational GNN, answer-conditioned QFormer, and scoring layers.
The text embedding model and rollout-generating search policy remain fixed.
Consequently, \textsc{TRACE} can be trained as a separate post-rollout selection module and then applied to candidate trajectories generated by different search policies or agent backbones without modifying those generators.

\section{Experiments}
\label{sec:experiments}
We organize our experiments around four research questions (RQs).
\textbf{RQ1: }
Can \textsc{TRACE} select better final answers from the same $K$ completed rollouts than answer-level heuristics and LLM-based generative aggregation?
\textbf{RQ2: }
Does a selector trained once per search setting transfer across rollout policies, model scales, and agent backbones without generator-specific fine-tuning?
\textbf{RQ3: }
Do occurrence preservation, cross-rollout evidence propagation, and answer-conditioned readout each contribute to the gains of \textsc{TRACE}?
\textbf{RQ4: }
Can a fixed selector exploit different rollout budgets, and can better selection reduce the number of rollouts needed for competitive accuracy?

\subsection{Experimental Setup}
\label{sec:experimental_setup}

\paragraph{Benchmarks and metrics.}
We evaluate \textsc{TRACE} in two search regimes.
Retrieval-based WebQA uses 3,125 questions from NQ~\citep{kwiatkowski2019natural}, HotpotQA~\citep{yang2018hotpotqa}, TriviaQA~\citep{joshi2017triviaqa}, PopQA~\citep{mallen2023memories}, 2WikiMultiHopQA~\citep{ho2020multihop}, MuSiQue~\citep{trivedi2022musique}, and Bamboogle~\citep{press2022compositionality}, and reports question-weighted exact match (EM).
Long-horizon online search uses BrowseComp-Plus~\citep{chen2025browsecompplus}, FRAMES~\citep{krishna2024frames}, and GAIA~\citep{mialon2023gaia}, with accuracy taken from the saved Qwen3-32B correctness judgments of the candidate pools.
Unless stated otherwise, the long-horizon aggregate averages the two rollout backbones within each benchmark and then the three benchmarks equally.

\paragraph{Rollout generators and transfer setting.}
WebQA candidate pools come from Base, SFT, and RL variants of Qwen2.5-7B and Qwen2.5-14B, giving six rollout policies with substantially different single-rollout accuracies and candidate distributions.
Long-horizon pools come from OpenResearcher-30B-A3B and gpt-oss-120B; a single long-horizon checkpoint is applied to both without agent-specific fine-tuning.

\paragraph{Selector training.}
For WebQA, we sample $K=16$ Qwen2.5-14B trajectories per question on the NQ and HotpotQA training sets (101,323 training and 5,309 validation questions).
For long-horizon search, we use existing 16-trajectory pools with correctness labels from the OpenResearcher dataset~\citep{li2026openresearcher} (2,655 training and 132 validation questions).
Both settings keep only questions whose pools contain both correct and incorrect answers.

\paragraph{Baselines.}
We compare three classes of post-rollout consolidation.
\emph{Answer-level heuristics} are majority voting~\citep{wang2023selfconsistency}, confidence-weighted majority voting, and Fewest Tools.
\emph{Generative aggregation} covers SolAgg, SummAgg, and AggAgent~\citep{lee2026aggagent}, all using Qwen3-32B as the aggregator regardless of the rollout generator.
Single-rollout accuracy serves as a reference; Appendix~\ref{app:baselines} gives the full protocols.

\paragraph{Implementation.}
We train for 3 epochs with AdamW (constant learning rate $3\times10^{-4}$, weight decay $10^{-4}$, dropout 0.1) and batch sizes of 256 (WebQA) and 4 (long-horizon) question graphs.
The type-specific projections, four-layer relation-specific GraphSAGE encoder, answer-conditioned QFormer, and scoring layers are optimized jointly, while the rollout policies and Qwen3-Embedding-8B stay frozen.
We hold out about 5\% of training questions for validation, select one checkpoint by validation performance, and keep it fixed in every rollout-budget sweep.
Appendix~\ref{app:implementation} details filtering, graph parsing, and evaluation.

\subsection{Main Results: Effective and Transferable Trajectory Selection (\textit{RQ1, RQ2})}
\label{sec:main_results}

\begin{table}[t]
\centering
\footnotesize
\setlength{\tabcolsep}{0pt}
\renewcommand{\arraystretch}{1.08}
\caption{WebQA EM (\%) at $K=16$ with Qwen2.5-14B Base and SFT rollouts.
SolAgg, SummAgg, and AggAgent use Qwen3-32B.
Overall is question-weighted within each rollout policy.
Bold marks the best result in each column.}
\label{tab:main_results}
\begin{tabular*}{\linewidth}{@{\extracolsep{\fill}}l@{\hspace{4pt}}*{7}{r@{\hspace{2.5pt}}r@{\hspace{4pt}}}r@{\hspace{3pt}}r@{}}
\toprule
 & \multicolumn{2}{c}{NQ} & \multicolumn{2}{c}{HotpotQA} & \multicolumn{2}{c}{TriviaQA} & \multicolumn{2}{c}{PopQA} & \multicolumn{2}{c}{2Wiki} & \multicolumn{2}{c}{MuSiQue} & \multicolumn{2}{c}{Bamboogle} & \multicolumn{2}{c}{Overall} \\
\cmidrule(lr){2-3}\cmidrule(lr){4-5}\cmidrule(lr){6-7}\cmidrule(lr){8-9}\cmidrule(lr){10-11}\cmidrule(lr){12-13}\cmidrule(lr){14-15}\cmidrule(lr){16-17}
Method & Base & SFT & Base & SFT & Base & SFT & Base & SFT & Base & SFT & Base & SFT & Base & SFT & Base & SFT \\
\midrule
Single rollout & 29.4 & 41.4 & 27.6 & 36.8 & 49.6 & 64.2 & 35.2 & 38.6 & 24.6 & 31.4 & 10.4 & 18.4 & 31.2 & 45.6 & 29.5 & 38.8 \\
\midrule
Majority Voting & 44.6 & 46.6 & 40.4 & 46.8 & 63.8 & 72.2 & 46.0 & 46.2 & 40.8 & 46.2 & 18.4 & 22.4 & 48.8 & 59.2 & 42.6 & 47.2 \\
Weighted Voting & 44.4 & 46.6 & 40.6 & 47.2 & 64.8 & 72.6 & 46.0 & 46.2 & 41.6 & 45.6 & 19.0 & 22.4 & 48.8 & 60.0 & 43.0 & 47.3 \\
Fewest Tools & 36.0 & 45.0 & 30.2 & 42.6 & 58.2 & 71.2 & 39.4 & 43.4 & 26.0 & 41.0 & 9.6 & 20.2 & 32.8 & 57.6 & 33.2 & 44.4 \\
\midrule
SolAgg & 43.2 & 43.4 & 42.6 & \textbf{49.0} & 65.0 & 72.4 & 45.0 & 46.4 & 45.8 & \textbf{48.6} & 19.6 & 25.0 & 52.8 & \textbf{60.8} & 43.9 & 48.0 \\
SummAgg & 41.2 & 41.6 & \textbf{44.2} & 48.0 & 62.2 & 71.6 & 43.6 & 44.0 & \textbf{48.0} & 47.2 & 21.2 & 24.6 & 51.2 & 58.4 & 43.7 & 46.7 \\
AggAgent & 41.0 & 39.6 & 37.8 & 44.0 & 63.6 & 68.0 & 40.4 & 42.2 & 44.8 & 43.2 & 17.0 & 21.0 & \textbf{60.0} & 59.2 & 41.5 & 43.6 \\
\midrule
\textbf{TRACE} & \textbf{47.4} & \textbf{48.4} & 43.2 & 48.2 & \textbf{65.6} & \textbf{73.2} & \textbf{47.6} & \textbf{49.2} & 44.8 & 47.2 & \textbf{21.4} & \textbf{26.0} & 51.2 & 60.0 & \textbf{45.2} & \textbf{49.2} \\
\bottomrule
\end{tabular*}
\end{table}

\paragraph{WebQA across rollout policies.}
Table~\ref{tab:main_results} compares selectors on Qwen2.5-14B Base and SFT pools, whose single-rollout accuracies differ substantially.
\textbf{Observation 1: \textsc{TRACE} improves final-answer selection across distinct WebQA rollout distributions.}
On Base pools, \textsc{TRACE} reaches 45.2\% overall EM, 2.6 and 2.2 points above majority and weighted voting, and improves over majority voting on all seven datasets.
On the stronger SFT pools, it reaches 49.2\% versus 47.2\% for majority voting.
Appendix~\ref{app:webqa_policies} shows that the same WebQA selector improves over both voting baselines on all six Qwen2.5-7B/14B Base, SFT, and RL policies.

\begin{table}[t]
\centering
\footnotesize
\setlength{\tabcolsep}{5pt}
\renewcommand{\arraystretch}{1.05}
\caption{Long-horizon accuracy (\%). Single rollout reports mean Pass@1 over 16 samples; other rows use $K=16$.
OR and OSS denote OpenResearcher-30B-A3B and gpt-oss-120B rollouts.
All generative aggregators use Qwen3-32B.
\textbf{Bold} marks the best result in each column.}
\vspace{1em}
\label{tab:browse_main_results}
\begin{tabular}{@{}lrrrrrrr@{}}
\toprule
& \multicolumn{2}{c}{BrowseComp-Plus} & \multicolumn{2}{c}{FRAMES} & \multicolumn{2}{c}{GAIA} & \\
\cmidrule(lr){2-3}\cmidrule(lr){4-5}\cmidrule(lr){6-7}
Method & OR & OSS & OR & OSS & OR & OSS & Avg. \\
\midrule
Single rollout & 38.3 & 45.2 & 73.1 & 84.9 & 55.4 & 61.7 & 59.7 \\
\midrule
Majority Voting & 64.5 & 66.0 & 88.4 & 90.4 & 66.0 & 77.7 & 75.5 \\
Weighted Voting & 64.5 & 70.2 & 88.4 & 90.6 & 67.0 & \textbf{79.6} & 76.7 \\
Fewest Tools & 64.6 & 72.7 & 86.0 & 89.0 & 51.5 & 76.7 & 73.4 \\
\midrule
SolAgg & 67.0 & 71.4 & 88.6 & 89.0 & 66.0 & 77.7 & 76.6 \\
SummAgg & \textbf{67.1} & 72.4 & 87.6 & 90.2 & 66.0 & 78.6 & 77.0 \\
AggAgent & 58.1 & 62.3 & 76.6 & 76.6 & 66.0 & 73.8 & 68.9 \\
\midrule
\textbf{TRACE} & \textbf{67.1} & \textbf{74.3} & \textbf{90.6} & \textbf{91.8} & \textbf{68.9} & 78.6 & \textbf{78.6} \\
\bottomrule
\end{tabular}
\end{table}

\paragraph{Long-horizon search across benchmarks and agent backbones.}
Table~\ref{tab:browse_main_results} applies one selector to three benchmarks and two rollout generators, where trajectories involve long multi-step browsing and cross-rollout sharing is captured through document identity rather than identical chunks.
\textbf{Observation 2: One long-horizon selector transfers across benchmarks and rollout generators.}
\textsc{TRACE} reaches 78.6\% average accuracy, 3.1 and 1.9 points above majority and weighted voting, and is best (including ties) in five of six settings.
The largest gain over majority voting is on gpt-oss-120B BrowseComp-Plus (66.0\% to 74.3\%).
Since one selector serves both generators, post-rollout selection can be decoupled from the model that generates the trajectories.

\textbf{Observation 3: Discriminative trajectory selection can outperform large generative aggregation.}
The generative baselines use Qwen3-32B to read all trajectories and synthesize a new response, whereas \textsc{TRACE} is a compact graph selector over frozen embeddings that returns an existing candidate.
Still, \textsc{TRACE} exceeds the strongest aggregator by 1.3/1.2 points on WebQA Base/SFT and by 1.6 points in the long-horizon setting, supporting post-rollout consolidation as selection when the pool already contains strong solutions.

\subsection{Mechanism Analysis (\textit{RQ3})}
\label{sec:ablations}

\begin{table}[t]
    \centering
    \small
    \setlength{\tabcolsep}{6pt}
    \caption{Mechanism ablations at $K=16$.
    WebQA reports EM (\%).
    }
    \vspace{1em}
    \label{tab:ablations}
    \begin{tabular}{@{}lrr@{}}
        \toprule
        Configuration & WebQA & Long-horizon avg. \\
        \midrule
        Full TRACE & \textbf{45.2} & \textbf{78.6} \\
        w/o cross-rollout communication & 43.9 & 72.7 \\
        w/o GNN & 43.7 & 75.5 \\
        Fixed-query readout & 44.7 & 75.6 \\
        \bottomrule
    \end{tabular}
\end{table}

Table~\ref{tab:ablations} tests whether the gains come from the relational mechanism of Section~\ref{sec:method}, holding candidate pools, embeddings, supervision, and optimization budget fixed.
\textbf{Observation 4: Cross-rollout shared evidence is particularly important for long-horizon selection.}
Removing only cross-rollout communication lowers WebQA EM from 45.2\% to 43.9\% and long-horizon accuracy from 78.6\% to 72.7\%, the largest drop among the main ablations, showing that evidence seen by other trajectories adds substantially beyond encoding each rollout independently.
Different trajectories often read different passages of the same source, and document identity links such evidence in a way that answer agreement or exact passage matching cannot.

\textbf{Observation 5: Graph propagation and answer-conditioned readout provide complementary gains.}
Removing the GNN gives 43.7/75.5 (WebQA/long-horizon), and replacing the answer-dependent query with a shared learned query gives 44.7/75.6.
This supports the two-stage design: first contextualize evidence across rollouts, then let each candidate read its own globally informed context.

Appendix~\ref{app:additional_ablations} reports further controls (WebQA/long-horizon).
Preserving occurrences matters: merging process occurrences, identical answers, or both gives 44.1/77.1, 44.9/76.3, and 44.4/76.4, indicating that each trajectory--answer occurrence should be kept rather than collapsing candidates that share evidence or answer strings.
The evidence-connection controls further validate the graph construction: removing content-identity links lowers WebQA EM to 43.8; in the long-horizon setting, no Doc or content links gives 71.1, content links alone 76.5, and Doc plus content 76.9, versus 78.6 for Doc links alone, supporting source identity as the relation for linking passages of the same document.
Replacing the QFormer with graph readout gives 44.7/75.8, and two-stage QFormer training gives 44.8/75.1, so jointly optimizing process encoding and readout performs best.

\subsection{Scaling and Efficiency (\textit{RQ4})}
\label{sec:rollout_scaling}

\begin{figure}[t]
    \centering
    \includegraphics[width=\linewidth]{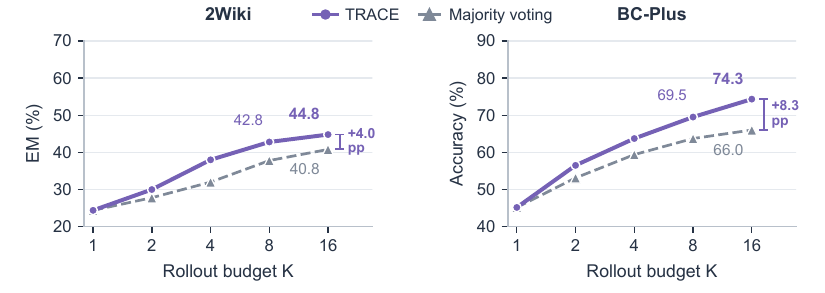}
    \caption{Final-answer performance with fixed \textsc{TRACE} checkpoints as the rollout budget increases.
    (a) 2Wiki EM with Qwen2.5-14B Base rollouts.
    (b) BrowseComp-Plus accuracy with gpt-oss-120B rollouts.
    Dashed curves show majority voting; brackets mark \textsc{TRACE} gains at $K=16$.}
    \label{fig:scaling}
\end{figure}

We apply the checkpoint trained at $K=16$ to pools of other sizes without retraining (Figure~\ref{fig:scaling}).
\textbf{Observation 6: A fixed selector generalizes across rollout budgets.}
\textsc{TRACE} stays above majority voting for every $K\geq2$, reaching 42.8\% and 44.8\% EM at $K=8$ and $K=16$ on 2Wiki, and rising from 69.5\% to 74.3\% on gpt-oss-120B BrowseComp-Plus.
Across all WebQA datasets, EM rises from 35.2\% at $K=2$ to 45.2\% at $K=16$ and 46.3\% at $K=32$, a budget unseen in training (Appendix~\ref{app:complete_results}).
Because the checkpoint is unchanged throughout the sweep, \textsc{TRACE} naturally operates over candidate sets that differ in size from those seen during training.

\begin{table}[!ht]
\centering
\small
\caption{Post-rollout runtime on the 3,125-question SFT WebQA pool. All LLM aggregators use Qwen3-32B. GPU-hours use H20s; relative cost is normalized to TRACE.}
\label{tab:efficiency}
\begin{tabular}{@{}lrr@{}}
\toprule
Method & GPU-hours $\downarrow$ & Relative cost \\
\midrule
\textbf{TRACE} & \textbf{0.271} & $1\times$ \\
SolAgg & 3.28 & $12\times$ \\
SummAgg & 32.96 & $122\times$ \\
AggAgent & 69.93 & $258\times$ \\
\bottomrule
\end{tabular}
\end{table}

\textbf{Observation 7: Better selection can substitute for substantially more rollout generation.}
On WebQA Base pools, \textsc{TRACE} at $K=8$ reaches 43.3\% EM, essentially matching majority voting at $K=64$ (43.6\%) with one eighth of the rollouts (Appendix~\ref{app:complete_results}).

\paragraph{Post-rollout processing cost.}
Processing all 3,125 SFT WebQA questions with \textsc{TRACE} takes only 0.271 H20 GPU-hours (16.2 minutes on one H20), including fresh text encoding, graph construction, and trajectory selection. As shown in Table~\ref{tab:efficiency}, this is substantially cheaper than all three Qwen3-32B generative aggregators: SolAgg, SummAgg, and AggAgent require 3.28, 32.96, and 69.93 GPU-hours, respectively, corresponding to approximately $12\times$, $122\times$, and $258\times$ the post-rollout cost of \textsc{TRACE}. Thus, even the fastest generative baseline requires more than an order of magnitude greater GPU compute, while the more elaborate aggregation pipelines incur two orders of magnitude higher cost. Appendix~\ref{app:efficiency} provides the runtime breakdown and timing scope.

\section{Conclusion}
\label{sec:conclusion}
We presented \textsc{TRACE}, a lightweight trajectory selector for parallel scaling of search agents. \textsc{TRACE} preserves retrieval occurrences, connects trajectories through shared evidence, and combines cross-rollout graph propagation with answer-conditioned readout to select an existing candidate without additional search or autoregressive aggregation. Across six WebQA rollout policies and long-horizon BrowseComp-Plus, FRAMES, and GAIA evaluations, one selector per search setting consistently improves over voting and remains competitive with or stronger than Qwen3-32B generative aggregation. These results show that post-rollout consolidation can be effectively decoupled from trajectory generation and handled by a lightweight, transferable selection module.

\label{main_text_end}
\FloatBarrier

\section*{AI Use Statement}
We used ChatGPT and Codex to create and edit code and to edit the manuscript for readability. We reviewed and verified all AI-assisted work and take responsibility for the final content of this work.

\section*{Reproducibility Statement}
Sections~\ref{sec:method} and~\ref{sec:experimental_setup} specify the model, objectives, and optimization settings. Appendix~\ref{app:implementation} provides implementation and evaluation details, Appendix~\ref{app:additional_ablations} gives additional ablations, and Appendices~\ref{app:complete_results}--\ref{app:efficiency} give supporting results, coverage analysis, and efficiency measurements.

All code, including the implementation, training configurations, and evaluation scripts, is available at \url{https://github.com/Jaasssoooonnnnn/TRACE}.

\bibliographystyle{iclr2027_conference}
\bibliography{references}

\clearpage
\appendix
\numberwithin{table}{section}
\numberwithin{figure}{section}
\numberwithin{equation}{section}
\section{Implementation and Evaluation Details}
\label{app:implementation}

\subsection{Candidate Pools and Metrics}
\label{app:data_protocols}
Training and validation use fixed question-level 5\% holdouts: deterministic for WebQA and stored with the long-horizon graphs. WebQA Single rollout scores unfiltered sample 0; long-horizon Single rollout and scaling at $K=1$ average all 16 original judged samples per question, counting failed or empty samples as wrong. At $K\geq2$, selection takes the first $K$ samples. WebQA retains trajectories with a final answer and retrieved evidence; long-horizon browsing retains nonempty final responses. Invalid samples are not replaced. Graph memberships and vote features are recomputed for each retained pool, and empty pools receive zero accuracy and F1 while remaining in the denominator.

WebQA EM and token F1 lowercase answers, remove punctuation and English articles, collapse whitespace, and take the best match over reference aliases. Long-horizon accuracy uses the saved Qwen3-32B judgments. WebQA aggregates weight each question equally. Long-horizon aggregates instead use $\frac{1}{3}\sum_d\frac{1}{2}\sum_b A_{d,b}$, where $A_{d,b}$ is accuracy for dataset $d$ and backbone $b$; each dataset and backbone receives equal weight. This same aggregation rule applies to the long-horizon ablations. The WebQA evaluation includes 500 questions from each dataset except Bamboogle, which contributes 125.

Long-horizon evaluation uses all 830 BrowseComp-Plus questions, all 103 questions in the GAIA text-only \texttt{gaia\_text} split of OpenResearcher/web-bench, and a sampled subset of 500 FRAMES questions. Question sets remain fixed across backbones, consolidation methods, ablations, and rollout budgets.

\paragraph{WebQA graph parsing.}
A retained rollout must contain a final answer and nonempty retrieved evidence. Each nonempty Search return is split into the ordered rank-1, rank-2, and rank-3 chunks. Rank markers are stored as edge types; the remaining chunk text retains its title and content. Identity normalizes Unicode to NFC and collapses whitespace, then compares text exactly. Each valid Search and each ranked return remains a separate occurrence. Empty returns create no retrieval nodes, and temporal edges follow the retained Search events. Shared text embeddings are lookup reuse, not merged graph nodes.

\paragraph{Browsing graph parsing.}
We retain nonempty final responses and their Search/Open/Find records. Every recorded Search receives a Subquery node. Open and Find results become Evidence nodes, with their original tool kind and event order. Parent-event and cursor references trace each observation to its Search ancestor; a chain with no such ancestor attaches to its own orphan Subquery instead of the most recent Search. The explicit temporal chain connects Subquery nodes in both directions. Evidence is attached to a Doc using the verified URL/view identity mapping. URL normalization resolves equivalent display and encoding forms and removes fragments, while preserving query information. Find and page-continuation views inherit their parent's source. Retained search-result views are scoped to their rollout, so a search listing cannot accidentally link unrelated webpages across trajectories. Each Evidence occurrence has one document membership. Only Subquery and Evidence states enter that rollout's attention context.

\subsection{Baselines}
\label{app:baselines}
Weighted Majority Voting sums confidence within each normalized-answer group. Confidence is estimated from the completed trajectory and final answer by its generating model, without reference answers. Fewest Tools counts Search calls for WebQA and Search/Open/Find calls for browsing. SolAgg integrates final solution texts; SummAgg first summarizes each interaction; AggAgent accesses candidates through solution, trajectory-search, and segment tools. Generative aggregation uses Qwen3-32B in the main and supplementary WebQA comparisons and for long-horizon browsing. SolAgg and SummAgg use temperature 1 and a 10,000-token output budget.

\paragraph{WebQA answer extraction.}
SolAgg retains the first complete answer tag. SummAgg retains tagged answers; when tags are absent, it accepts an explicit first-line declaration beginning with ``The correct answer is:'', ``The answer is:'', ``Final answer:'', or ``Answer:'', removing Markdown bold markers and a trailing period. For AggAgent, we take the first nonempty line of an accepted answer. For completed runs without an accepted answer, we use the first saved \texttt{finish.solution} verbatim; runs without a finish call and terminated runs retain empty answers. These post-hoc rules use no reference answers to choose the extracted text. Long-horizon accuracy uses the saved correctness judgments for final responses.

\subsection{Graph Statistics and Counting Units}
\label{app:graph_statistics}
We count the retained $K=16$ evaluation pools for Qwen2.5-14B Base on WebQA and gpt-oss-120B on BrowseComp-Plus: 3,125 and 830 question graphs, respectively. Counts use the retained graph metadata and document identities described above. WebQA has 39,433 retained rollouts and 188,889 Evidence (chunk) occurrences; browsing has 9,509 retained rollouts and 97,781 Evidence occurrences.

For a content or document group $g$, let $n_{g,r}$ be its number of occurrences in rollout $r$. Its number of cross-rollout pairs is $\sum_{r<s}n_{g,r}n_{g,s}$, so each unordered pair is counted once and same-rollout repetitions are excluded. A shared group is a distinct content or source identity present in at least two rollouts. Table~\ref{tab:graph_statistics} reports unweighted means over question graphs, including graphs with zero shared groups. Medians and 90th percentiles show the distribution of pair counts.

\begin{table}[htbp]
\centering
\small
\caption{Cross-rollout sharing at $K=16$. Groups and occurrence pairs are counted within each question graph. ``Graphs'' is the percentage containing at least one such pair.}
\label{tab:graph_statistics}
\begin{tabular}{@{}llrrrrr@{}}
\toprule
Task & Matching rule & Groups & \multicolumn{3}{c}{Occurrence pairs} & Graphs (\%) \\
\cmidrule(lr){4-6}
 & & Mean & Mean & Median & P90 & \\
\midrule
WebQA & Identical chunk & 7.6 & 243.6 & 228 & 430 & 99.9 \\
Browsing & Identical evidence & 15.4 & 462.0 & 314 & 1,080 & 99.0 \\
Browsing & Same document & 6.5 & 1,797.6 & 1,142.5 & 4,224 & 99.0 \\
\bottomrule
\end{tabular}
\end{table}

The graph representation determines how these pairs become messages. WebQA instantiates each identical-content pair in both directions: an average of 487.2 cross-rollout and 14.4 within-rollout directed edges per graph. Browsing uses document incidence edges, not an Evidence clique: its 117.8 Evidence occurrences per graph give 235.6 directed Evidence--Doc edges, providing two-hop paths between same-source observations. Of 1,491,965 cross-rollout same-document pairs, 1,108,528 (74.3\%) have different content identities. A document observed $m$ times requires $2m$ directed incidence edges, compared with $m(m-1)$ for a pairwise same-source clique.

\subsection{Encoder and Readout Details}
\label{app:encoder_details}
Frozen Qwen3-Embedding-8B~\citep{zhang2025qwen3embedding} supplies text embeddings $\mathbf{x}_v$. The embedding pipeline pools the last non-padding token and applies $\ell_2$ normalization. A trainable encoder $\phi_\nu(\mathbf{x})=\operatorname{GELU}(\operatorname{LN}(\mathbf{W}_\nu\mathbf{x}+\mathbf{b}_\nu))$ maps each text-bearing node type $\nu$ to dimension $d$, giving $\mathbf{h}_v^{(0)}=\phi_{\operatorname{type}(v)}(\mathbf{x}_v)$. Answer representations additionally incorporate answer frequency:
\begin{equation}
    \mathbf{h}_{a_i}^{(0)}=\phi_{\mathrm{A}}(\mathbf{x}_{a_i})+\mathbf{W}_{\mathrm{vote}}\mathbf{f}_i,
    \qquad
    \mathbf{f}_i=\big[\log(1+c_i),\;c_i/R_q\big]^{\top},
    \label{eq:app_answer_initialization}
\end{equation}
where $c_i$ counts retained candidates with the same normalized answer. Doc nodes share a learned initial vector $\mathbf{d}_0$, initialized to zero, and obtain document-specific information through their neighbors. Orphan Subqueries have a zero text-embedding input and receive an additional learned indicator vector, initialized to zero; no synthetic query text is embedded.

We use relation-specific GraphSAGE~\citep{hamilton2017inductive} to propagate process information. For incoming relation types $r$ and their neighbor sets $\mathcal{N}_r(v)$, layer $\ell\in\{0,\ldots,L-1\}$ computes
\begin{equation}
\begin{aligned}
    \mathbf{m}_v^{(\ell)}
    &=\sum_r\left(\mathbf{W}_r^{(\ell)}
      \operatorname{Mean}_{u\in\mathcal{N}_r(v)}\mathbf{h}_u^{(\ell)}
      +\mathbf{b}_r^{(\ell)}\right),\\
    \mathbf{h}_v^{(\ell+1)}
    &=\operatorname{LN}\!\left(\mathbf{h}_v^{(\ell)}+
      \operatorname{Dropout}\!\left(\operatorname{GELU}(\mathbf{m}_v^{(\ell)})\right)\right).
\end{aligned}
\label{eq:app_process_message_passing}
\end{equation}
Updates are synchronous, with node self-information retained by the residual. An empty relation neighborhood has a zero mean; the relation transformation still includes its bias. We use $L=4$ layers with $d=256$. Subquery and evidence states are updated in both variants, along with Doc states in long-horizon browsing. Query and Answer states receive no graph messages. Cross-trajectory information travels directly between matching WebQA chunks or along the two-hop Evidence--Doc--Evidence path before answer readout.

The QFormer has one block with four heads of width 64. Each head uses the answer state as its query and the rollout's Subquery/Evidence states as keys and values. Head outputs are concatenated and projected, followed by a $256\!\rightarrow\!1024\!\rightarrow\!256$ GELU feed-forward network. Both sublayers use dropout 0.1, residual connections, and LayerNorm. If $\mathcal{C}_i$ is empty, the attention sum is zero and the residual, feed-forward, and normalization operations still apply. The question and answer readouts $\rho_{\mathrm{Q}}$ and $\rho_{\mathrm{A}}$ are separate affine $256\!\rightarrow\!256$ maps. Scoring uses $\gamma=\min(\exp(\eta),100)$ with $\eta$ initialized to $\log 10$.

\paragraph{Loss definitions.}
With the positive and negative candidate sets defined in Section~\ref{sec:selector_training}, the three terms are
\begin{equation}
\begin{aligned}
    \ell_{\mathrm{BCE}}(i)
    &=-w_+y_i\log\sigma(\mathrm{score}_i)
      -(1-y_i)\log\!\left(1-\sigma(\mathrm{score}_i)\right),\\
    \ell_{\mathrm{list}}(q)
    &=\log\sum_{i\in\mathcal{I}_q}e^{\mathrm{score}_i}
      -\log\sum_{i\in\mathcal{P}_q}e^{\mathrm{score}_i},\\
    \ell_{\mathrm{hard}}(q)
    &=\operatorname{softplus}\!\left(
      \max_{i\in\mathcal{N}_q}\mathrm{score}_i-\max_{i\in\mathcal{P}_q}\mathrm{score}_i\right),
\end{aligned}
\label{eq:selector_losses}
\end{equation}
Here $\sigma$ is the sigmoid and $w_+$ is the ratio of negative to positive training candidates. BCE is averaged over candidates and each ranking term over its eligible questions.

\paragraph{Loss eligibility and empty pools.}
WebQA averages listwise terms over questions with at least one correct candidate; long-horizon browsing uses pools containing both correct and incorrect candidates. Hard ranking applies only to mixed pools in both settings. Empty pools have no selected candidate and count as incorrect, as specified in Appendix~\ref{app:data_protocols}.

\FloatBarrier
\section{Additional Ablations}
\label{app:additional_ablations}

\paragraph{Protocol.}
All comparisons use $K=16$ with the same question splits, frozen text embeddings, correctness supervision, evaluation pools, and optimization budget. The controls that remove cross-rollout graph messages, bypass the GNN, or use a fixed attention query in Table~\ref{tab:ablations} retain the full model's answer-frequency features. The fixed query adds a 256-dimensional learned vector; bypassing the GNN leaves the type encoders, attention readout, and scoring layers trainable. Readout membership remains local to each trajectory in these three controls.

\begin{table}[!htbp]
    \centering
    \small
    \setlength{\tabcolsep}{6pt}
    \caption{Additional design comparisons at $K=16$. Columns report WebQA EM and long-horizon average accuracy (\%), using the same aggregation as Table~\ref{tab:ablations}. Dashes denote unevaluated comparisons.}
    \label{tab:additional_ablations}
    \begin{tabular}{@{}lrr@{}}
        \toprule
        Configuration & WebQA & Long-horizon avg. \\
        \midrule
        Full TRACE & \textbf{45.2} & \textbf{78.6} \\
        \midrule
        \multicolumn{3}{@{}l}{\emph{Occurrence representation}} \\
        Merge identical evidence occurrences & 44.1 & 77.1 \\
        Merge identical answers & 44.9 & 76.3 \\
        Merge both & 44.4 & 76.4 \\
        \midrule
        \multicolumn{3}{@{}l}{\emph{Answer readout and training}} \\
        Graph readout (no attention) & 44.7 & 75.8 \\
        Two-stage readout training & 44.8 & 75.1 \\
        \midrule
        \multicolumn{3}{@{}l}{\emph{Evidence connections}} \\
        No content-identity links & 43.8 & --- \\
        No Doc or content-identity links & --- & 71.1 \\
        Content-identity links only & --- & 76.5 \\
        Doc and content-identity links & --- & 76.9 \\
        \bottomrule
    \end{tabular}
\end{table}

\paragraph{Answer-merging controls.}
Merged answers use canonical-answer groups, the first member's answer embedding, and the union of their trajectories' Subquery/Evidence memberships. Group labels average member correctness. In the browsing control, BCE uses the resulting soft labels; listwise and hard-ranking losses treat groups with mean correctness greater than 0.5 as positive and the remaining groups as negative. Merging preserves the occurrence graph's question split, process nodes, and process edges. The browsing readout and answer-merging controls use the full model's frozen embeddings and source identities.

\paragraph{Occurrence representation and evidence connections.}
The process-merging control combines identical-content evidence occurrences and their incident retrieval connections. Answer merging changes the evaluation unit: identical answers share a readout over the union of their trajectories' contexts, as defined above. On long-horizon browsing, merging process occurrences, answers, or both gives 77.1\%, 76.3\%, and 76.4\%, respectively, compared with 78.6\% for Full TRACE. The corresponding WebQA results are 44.1\%, 44.9\%, and 44.4\%, compared with 45.2\%. These comparisons support preserving each occurrence's process context when ranking trajectories.

The connection controls change which identities link evidence. On WebQA, removing all content-identity links gives 43.8\% EM. In long-horizon browsing, removing both Doc and content links gives 71.1\% average accuracy, 7.4 points below Full TRACE. Direct content links alone achieve 76.5\%, while combining Doc and content links gives 76.9\%. Doc-based TRACE reaches 78.6\%, supporting source identity as a useful relation between different passages while preserving their query and tool contexts. These broader interventions also remove within-rollout matches or document aggregation, unlike the ablation of cross-rollout graph messages in the main text.

\paragraph{Readout architecture and training.}
Graph readout replaces cross-attention with process-to-answer GNN messages. Two-stage training first fits this graph selector, then freezes its encoders, GNN, and scoring layers while training a residual attention readout. Their long-horizon averages are 75.8\% and 75.1\%, respectively, compared with 78.6\% when TRACE jointly trains its process encoder and attention readout. The resulting gains of 2.7 and 3.4 points support jointly optimizing process encoding and answer-conditioned readout. The fixed-query control in Table~\ref{tab:ablations} more directly tests answer conditioning by retaining attention and changing its query.

\FloatBarrier
\section{Selection under Different Rollout Budgets}
\label{app:complete_results}

Table~\ref{tab:app_selector_budget} reports aggregate fixed-checkpoint budget results, complementing the individual dataset--backbone curves in Figure~\ref{fig:scaling}. WebQA uses Qwen2.5-14B Base rollouts. Long-horizon results weight the six dataset--backbone combinations equally, with the same 500-question FRAMES subset at every budget. For long-horizon $K=1$, we report mean Pass@1; $K\geq2$ uses the first $K$ samples.
\begin{table}[htbp]
\centering
\small
\caption{Budget-scaling results (\%). WebQA: TRACE EM/F1 on Qwen2.5-14B Base rollouts through $K=32$. Long-horizon: accuracy through $K=16$.}
\label{tab:app_selector_budget}
\begin{tabular}{@{}rrrrrr@{}}
\toprule
& \multicolumn{2}{c}{WebQA} & \multicolumn{3}{c}{Long-horizon avg.} \\
\cmidrule(lr){2-3}\cmidrule(l){4-6}
$K$ & EM & F1 & TRACE & Majority & Pass@$K$ \\
\midrule
1 & 28.5 & 36.4 & 59.7 & 59.7 & 59.7 \\
2 & 35.2 & 43.9 & 67.8 & 65.5 & 72.2 \\
4 & 40.2 & 48.8 & 71.5 & 69.7 & 80.1 \\
8 & 43.3 & 51.8 & 75.3 & 73.8 & 86.1 \\
16 & 45.2 & 53.4 & 78.6 & 75.5 & 90.9 \\
32 & 46.3 & 54.6 & --- & --- & --- \\
\bottomrule
\end{tabular}
\end{table}

\FloatBarrier
\section{Policy-Sampling Results}
\label{app:policy_scaling_tables}

Table~\ref{tab:app_policy_scaling} reports the Pass@$K$ curves discussed in Appendix~\ref{sec:parallel_search}. For each question, let $c_q$ denote the number of correct trajectories among 64 samples. We estimate Pass@$K$ as $1-\binom{64-c_q}{K}/\binom{64}{K}$ and average over questions, taking the numerator to be zero when $64-c_q<K$. Thus Pass@1 averages correctness over all 64 samples per question, whereas the WebQA single-rollout references evaluate sample 0. SFT coverage uses all generated samples before selector filtering; normalized empty predictions count as incorrect.

The Qwen2.5-7B and Qwen2.5-14B SFT policies are trained with full-parameter updates for three epochs on 3,083 successful NQ/HotpotQA search trajectories generated by Qwen3.6-35B-A3B. Each policy is evaluated on 64 samples for each of the same 3,125 WebQA questions. In the table, Q2.5 and Q3 abbreviate Qwen2.5 and Qwen3.
\begin{table}[htbp]
\centering
\footnotesize
\setlength{\tabcolsep}{3pt}
\caption{Question-weighted Pass@$K$ (\%) for the policies in the parallel-search study.}
\label{tab:app_policy_scaling}
\begin{tabular}{@{}lrrrrrrr@{}}
\toprule
Policy & 1 & 2 & 4 & 8 & 16 & 32 & 64 \\
\midrule
Q2.5-7B Base & 24.9 & 35.3 & 44.8 & 52.6 & 58.6 & 63.4 & 67.4 \\
Q2.5-14B & 30.5 & 40.1 & 48.0 & 54.3 & 59.3 & 63.6 & 67.3 \\
SearchR1 RL & 43.7 & 46.5 & 48.8 & 50.6 & 52.1 & 53.3 & 54.2 \\
Q2.5-32B & 30.5 & 39.7 & 47.6 & 54.2 & 59.4 & 63.6 & 67.4 \\
Q3-32B & 36.1 & 43.8 & 50.0 & 54.7 & 58.5 & 61.7 & 64.2 \\
SFT (7B) & 36.3 & 43.4 & 49.4 & 54.4 & 58.5 & 62.1 & 65.3 \\
SFT (14B) & 38.6 & 44.4 & 49.3 & 53.5 & 57.4 & 60.9 & 64.0 \\
\bottomrule
\end{tabular}
\end{table}

\FloatBarrier
\section{Additional WebQA Rollout Policies}
\label{app:webqa_policies}

Table~\ref{tab:app_webqa_policies} reports all six WebQA rollout policies with Qwen3-32B generative aggregation. The main WebQA table uses Qwen3-32B aggregators; Appendix~\ref{app:webqa_32b} gives the comparison by dataset. TRACE improves over majority and weighted voting across all six policies. Table~\ref{tab:app_webqa_policies} presents these voting comparisons as transfer across rollout generators.

\begin{table}[!htbp]
\centering
\footnotesize
\setlength{\tabcolsep}{2.5pt}
\renewcommand{\arraystretch}{1.05}
\caption{WebQA EM (\%) at $K=16$ across six rollout policies, weighted by question count over the seven datasets. SolAgg, SummAgg, and AggAgent use Qwen3-32B. Bold marks the best result in each column.}
\label{tab:app_webqa_policies}
\begin{tabular*}{\linewidth}{@{\extracolsep{\fill}}lrrrrrr@{}}
\toprule
& \multicolumn{3}{c}{Qwen2.5-7B rollouts} & \multicolumn{3}{c}{Qwen2.5-14B rollouts} \\
\cmidrule(lr){2-4}\cmidrule(l){5-7}
Method & Base & SFT & RL & Base & SFT & RL \\
\midrule
Single rollout & 24.6 & 36.6 & 44.0 & 29.5 & 38.8 & 46.9 \\
\midrule
Majority Voting & 38.1 & 46.0 & 45.1 & 42.6 & 47.2 & 47.6 \\
Weighted Voting & 38.5 & 46.1 & 45.3 & 43.0 & 47.3 & 47.7 \\
Fewest Tools & 27.2 & 42.0 & 45.3 & 33.2 & 44.4 & 47.1 \\
\midrule
SolAgg & 42.4 & 48.1 & \textbf{46.5} & 43.9 & 48.0 & \textbf{48.5} \\
SummAgg & \textbf{42.9} & 46.8 & 44.9 & 43.7 & 46.7 & 46.4 \\
AggAgent & 40.7 & 43.3 & 44.2 & 41.5 & 43.6 & 46.0 \\
\midrule
\textbf{TRACE} & 41.2 & \textbf{48.3} & 45.8 & \textbf{45.2} & \textbf{49.2} & 47.9 \\
\bottomrule
\end{tabular*}
\end{table}

The RL pools contain much more repetitive search evidence. We measure overlap using the Jaccard similarity between the retrieved-chunk sets of two retained trajectories, averaging over pairs within each question and then over questions with at least two retained trajectories. Mean overlap reaches 0.87 and 0.83 for 7B and 14B RL, compared with 0.36 and 0.50 for Base and 0.56 and 0.60 for SFT. The difference persists on matched questions where both policies have correct and incorrect candidates. Repetition creates many graph links, but these links connect fewer distinct evidence contexts. This reduced complementarity is consistent with the smaller gains over voting on RL rollouts: 0.7 points at 7B and 0.3 points at 14B.

For 7B Base, candidate availability is a more immediate constraint. Of the first 16 samples per question, 27.2\% are excluded because they lack a nonempty answer or retrieved evidence, compared with 21.1\% for 14B Base. Zero-search trajectories are especially common: 18.1\% at 7B versus 6.0\% at 14B. Some already answer correctly, but cannot enter the graph under the evidence requirement. Filtering reduces the fraction of questions with a correct candidate from 59.0\% to 56.5\% at 7B, compared with 59.2\% to 58.2\% at 14B. The resulting 7B pool contains 11.6 candidates per question on average. TRACE still gains 3.0 points over majority voting within this smaller pool, recovering useful distinctions among the remaining trajectories.

\FloatBarrier
\section{Qwen3-32B Aggregation on WebQA}
\label{app:webqa_32b}

Table~\ref{tab:webqa_32b} evaluates Qwen3-32B aggregation on the same Qwen2.5-14B Base and SFT pools as Table~\ref{tab:main_results}. Replacing Qwen2.5-14B with Qwen3-32B improves all three generative methods on these pools. On SFT rollouts, SolAgg rises from 33.9\% to 48.0\% Overall EM, SummAgg from 17.7\% to 46.7\%, and AggAgent from 29.3\% to 43.6\%. TRACE reaches 49.2\%, retaining the highest Overall EM in this comparison without an additional autoregressive aggregation stage.

\begin{table}[!htbp]
\centering
\footnotesize
\setlength{\tabcolsep}{0pt}
\renewcommand{\arraystretch}{1.08}
\caption{WebQA EM (\%) with Qwen3-32B generative aggregation at $K=16$ with Qwen2.5-14B Base and SFT rollouts. SolAgg, SummAgg, and AggAgent use Qwen3-32B. Overall is question-weighted within each rollout policy. Bold marks the best result in each column.}
\label{tab:webqa_32b}
\begin{tabular*}{\linewidth}{@{\extracolsep{\fill}}l@{\hspace{4pt}}*{7}{r@{\hspace{2.5pt}}r@{\hspace{4pt}}}r@{\hspace{3pt}}r@{}}
\toprule
 & \multicolumn{2}{c}{NQ} & \multicolumn{2}{c}{HotpotQA} & \multicolumn{2}{c}{TriviaQA} & \multicolumn{2}{c}{PopQA} & \multicolumn{2}{c}{2Wiki} & \multicolumn{2}{c}{MuSiQue} & \multicolumn{2}{c}{Bamboogle} & \multicolumn{2}{c}{Overall} \\
\cmidrule(lr){2-3}\cmidrule(lr){4-5}\cmidrule(lr){6-7}\cmidrule(lr){8-9}\cmidrule(lr){10-11}\cmidrule(lr){12-13}\cmidrule(lr){14-15}\cmidrule(lr){16-17}
Method & Base & SFT & Base & SFT & Base & SFT & Base & SFT & Base & SFT & Base & SFT & Base & SFT & Base & SFT \\
\midrule
Single rollout & 29.4 & 41.4 & 27.6 & 36.8 & 49.6 & 64.2 & 35.2 & 38.6 & 24.6 & 31.4 & 10.4 & 18.4 & 31.2 & 45.6 & 29.5 & 38.8 \\
\midrule
Majority Voting & 44.6 & 46.6 & 40.4 & 46.8 & 63.8 & 72.2 & 46.0 & 46.2 & 40.8 & 46.2 & 18.4 & 22.4 & 48.8 & 59.2 & 42.6 & 47.2 \\
Weighted Voting & 44.4 & 46.6 & 40.6 & 47.2 & 64.8 & 72.6 & 46.0 & 46.2 & 41.6 & 45.6 & 19.0 & 22.4 & 48.8 & 60.0 & 43.0 & 47.3 \\
Fewest Tools & 36.0 & 45.0 & 30.2 & 42.6 & 58.2 & 71.2 & 39.4 & 43.4 & 26.0 & 41.0 & 9.6 & 20.2 & 32.8 & 57.6 & 33.2 & 44.4 \\
\midrule
SolAgg & 43.2 & 43.4 & 42.6 & \textbf{49.0} & 65.0 & 72.4 & 45.0 & 46.4 & 45.8 & \textbf{48.6} & 19.6 & 25.0 & 52.8 & \textbf{60.8} & 43.9 & 48.0 \\
SummAgg & 41.2 & 41.6 & \textbf{44.2} & 48.0 & 62.2 & 71.6 & 43.6 & 44.0 & \textbf{48.0} & 47.2 & 21.2 & 24.6 & 51.2 & 58.4 & 43.7 & 46.7 \\
AggAgent & 41.0 & 39.6 & 37.8 & 44.0 & 63.6 & 68.0 & 40.4 & 42.2 & 44.8 & 43.2 & 17.0 & 21.0 & \textbf{60.0} & 59.2 & 41.5 & 43.6 \\
\midrule
\textbf{TRACE} & \textbf{47.4} & \textbf{48.4} & 43.2 & 48.2 & \textbf{65.6} & \textbf{73.2} & \textbf{47.6} & \textbf{49.2} & 44.8 & 47.2 & \textbf{21.4} & \textbf{26.0} & 51.2 & 60.0 & \textbf{45.2} & \textbf{49.2} \\
\bottomrule
\end{tabular*}
\end{table}

\FloatBarrier
\section{Candidate Coverage and Selection}
\label{sec:parallel_search}

\subsection{Parallel Search and Evaluation}
\label{sec:search_evaluation}

We distinguish the opportunities created by repeated search from the ability to select a successful trajectory~\citep{brown2024monkeys}. For a question $q$ and a fixed search policy $\pi$, let $\mathcal T_K=(\tau_1,\ldots,\tau_K)$ contain $K$ completed rollouts, with final answers $a_i$ and correctness labels $y_i\in\{0,1\}$. Define $E_K=\{\max_i y_i=1\}$ as the event that the pool contains a correct answer. Candidate coverage is $C_K(\pi)=\Pr(E_K)$, or Pass@$K$. A selector $S$ returns an index $\hat\imath=S(q,\mathcal T_K)$ without observing the correctness labels. Its final-answer accuracy satisfies, for $C_K(\pi)>0$,
\begin{equation}
    A_K(\pi,S)
    = \Pr(y_{\hat\imath}=1)
    = C_K(\pi)\,\Pr(y_{\hat\imath}=1\mid E_K).
    \label{eq:coverage_selection}
\end{equation}
Probabilities range over questions and sampled rollouts. This decomposition separates candidate coverage from selection success conditional on a correct answer being available. On fixed candidate pools, oracle accuracy is the fraction of questions containing a correct candidate. Comparisons with this oracle concern selection from the same pools, rather than methods that may generate new answers.

\subsection{Policy Choice under Parallel Search}
\label{sec:policy_scaling}

Table~\ref{tab:app_policy_scaling} compares Base, full-SFT, and Search-R1 RL~\citep{jin2025searchr1} with the same Qwen2.5-7B backbone on seven WebQA datasets. Results are weighted by question count; Appendix~\ref{app:policy_scaling_tables} gives the SFT protocol and full coverage results. The policy ordering changes with the sampling budget. Equation~\ref{eq:coverage_selection} further shows that coverage must be paired with successful selection, which we learn from completed rollouts while holding their generating policy fixed.

\subsection{A Probabilistic View of the Coverage--Voting Gap}
\label{sec:selection_gap}

For a fixed policy $\pi$, assume rollouts are i.i.d.\ conditional on $q$ over finitely many normalized answer classes. Let $\mathcal A_q^+$ denote the correct classes, $p_q$ their total probability, and $a_q^\star$ the unique modal class. Majority voting $S_{\mathrm{MV}}$ selects the most frequent class. Independence and the law of large numbers give
\begin{equation}
\begin{aligned}
    C_K(\pi)&=\mathbb E_q\!\left[1-(1-p_q)^K\right]
        \xrightarrow{K\to\infty}\Pr\nolimits_q(p_q>0),\\
    A_K(\pi,S_{\mathrm{MV}})&\xrightarrow{K\to\infty}
        \Pr\nolimits_q(a_q^\star\in\mathcal A_q^+).
\end{aligned}
\label{eq:coverage_voting_limits}
\end{equation}
The limiting gap is therefore $\Pr_q(p_q>0,\,a_q^\star\notin\mathcal A_q^+)$: sampling eventually finds any reachable correct answer, whereas voting converges to the most probable class, even if incorrect. For probabilities $(0.2,0.5,0.3)$ with only the first class correct, coverage tends to one and voting accuracy to zero. This gap requires no dependence between rollouts.

\FloatBarrier
\section{Efficiency Measurements}
\label{app:efficiency}

We measure post-rollout runtime on the full 3,125-question WebQA evaluation with Qwen2.5-14B SFT rollouts and NVIDIA H20 GPUs. The first 16 samples provide 35,503 retained candidates across 2,939 nonempty pools; all questions remain in the denominator. Table~\ref{tab:efficiency} compares H20 GPU-hours for TRACE and the Qwen3-32B generative aggregators. Rollout generation and correctness judging are excluded.

TRACE profiling covers text encoding, graph construction, and answer selection from completed candidate texts. The 68,664 distinct node texts are encoded once each with Qwen3-Embedding-8B, using BF16, batches of 16, a 4,096-token limit, and 4,096-dimensional outputs. The selector scores graphs in batches of 256. GPU timings use CUDA synchronization. Table~\ref{tab:efficiency_breakdown} includes the frozen encoder, which accounts for most of the inference time.

\begin{table}[!htbp]
\centering
\small
\caption{Measured TRACE runtime on one H20 for the full 3,125-question pool. Text embeddings are recomputed, not loaded from a previous run.}
\label{tab:efficiency_breakdown}
\begin{tabular}{@{}lr@{}}
\toprule
Stage & Seconds \\
\midrule
Read and parse candidate texts & 23.6 \\
Fresh text encoding & 905.2 \\
Assemble tensors and construct graphs & 38.2 \\
Score and extract answers & 7.2 \\
\midrule
Total & 974.2 \\
\bottomrule
\end{tabular}
\end{table}

Model initialization adds 52.8 seconds to the reported runtime. When graph tensors are reused, batched selector inference takes a median of 1.09 seconds over three measurements, including batching and device transfer.

The Qwen3-32B SummAgg evaluation generates 35,503 trajectory summaries and 2,939 final integrations. With two H20s using tensor parallelism, the complete pipeline takes 59,322 seconds (16.5 hours). This includes summary generation, final integration, auxiliary confidence scoring, and client scheduling, and excludes rollout generation and correctness judging.

GPU-hours in Table~\ref{tab:efficiency} sum elapsed time multiplied by the number of H20 GPUs used in each interval. SolAgg takes approximately 98.5 minutes on two H20s (3.28 GPU-hours), reconstructed from the first non-pilot request to the final confidence output; the two pilot questions are reused. SummAgg uses 32.96 GPU-hours. AggAgent uses an estimated 69.93 GPU-hours, integrating its initial two-GPU allocation and later four-GPU allocation through the terminal cutoff. This deployment estimate includes restarts and abnormal waits, with two unfinished questions counted as failures. SolAgg and SummAgg timings include auxiliary confidence scoring; concurrent request durations are not summed as batch runtime.

A three-epoch selector fit on 101,323 training questions, with validation on 5,309 questions, takes 4.17 minutes on one H20 using precomputed embeddings; including data loading and setup, it takes 6.42 minutes.

\end{document}